# Integrative CAM: Adaptive Layer Fusion for Comprehensive Interpretation of CNNs

Aniket K Singh[a], Debasis Chaudhuri[a,*], Manish P Singh[b] and Samiran Chattopadhyay[a]

[a]*CSE Department, Techno India University, Sector – V, Salt Lake, Kolkata – 91, West Bengal, India*

[b]*DRDO Young Scientist Laboratory – CT, IIT-M Research Park, Taramani, Chennai – 600113 Tamil Nadu, India*

**Abstract**

With the growing demand for interpretable deep learning models, this paper introduces Integrative CAM, an advanced Class Activation Mapping (CAM) technique aimed at providing a holistic view of feature importance across Convolutional Neural Networks (CNNs). Traditional gradient-based CAM methods, such as Grad-CAM and Grad-CAM++, primarily use final layer activations to highlight regions of interest, often neglecting critical features derived from intermediate layers. Integrative CAM addresses this limitation by fusing insights across all network layers, leveraging both gradient and activation scores to adaptively weight layer contributions, thus yielding a comprehensive interpretation of the model's internal representation. Our approach includes a novel bias term in the saliency map calculation, a factor frequently omitted in existing CAM techniques, but essential for capturing a more complete feature importance landscape, as modern CNNs rely on both weighted activations and biases to make predictions. Additionally, we generalize the alpha term from Grad-CAM++ to apply to any smooth function, expanding CAM applicability across a wider range of models. Through extensive experiments on diverse and complex datasets, Integrative CAM demonstrates superior fidelity in feature importance mapping, effectively enhancing interpretability for intricate fusion scenarios and complex decision-making tasks. By advancing interpretability methods to capture multi-layered model insights, Integrative CAM provides a valuable tool for fusion-driven applications, promoting the trustworthy and insightful deployment of deep learning models.

## 1. Introduction

The rapid development of machine learning, particularly through deep neural networks, has revolutionized artificial intelligence (AI), enabling advancements across a wide array of real-world applications, from autonomous systems to complex decision support systems [1]. Despite the exceptional capabilities of deep learning models in domains like object detection, speech recognition, and machine translation, these models have shifted away from the explainable, logic-driven approaches of traditional AI. In contrast to earlier systems, which could articulate their decision-making processes through transparent, traceable inference steps, modern AI systems often function as "black boxes," making it difficult to discern the reasoning behind their outputs [2, 3]. This lack of transparency is especially problematic in critical fields such as security, healthcare, and autonomous navigation, where understanding the basis for AI decisions is essential to ensure trust and facilitate human collaboration [4].

---

*Corresponding author
*Email addresses:* deba_chaudhuri@yahoo.co.in

In response to this challenge, researchers have explored a variety of methods to enhance the explainability of deep learning models. One prominent strategy involves training secondary models to provide rationales for the predictions of the primary models, thereby explaining the underlying decision-making processes [5, 6] Another approach involves manipulating the input data to examine the resulting changes in the model's outputs, to identify which features drive the model's decisions [7, 8]. Despite these promising directions, achieving full transparency and interpretability remains a complex and ongoing challenge, with solutions varying greatly across different problem domains [9, 10]. In areas where AI performance does not yet surpass human capabilities, transparency can help identify limitations and guide improvements [11]. As AI systems reach human performance, fostering user trust through clear, understandable explanations becomes critical [12]. Furthermore, in scenarios where AI outpaces human capabilities, explanations assume the role of machine teaching, empowering humans to make better decisions through AI-guided insights [13]. Ultimately, as AI systems become more integrated into society, it is essential to facilitate their adoption through greater interpretability, ensuring their meaningful use in diverse real-world contexts.

The trade-off between model accuracy and interpretability has long been a defining issue in AI development. While traditional rule-based systems are highly interpretable, they often lack accuracy or robustness [14]. Conversely, decomposable pipelines, where each stage is meticulously crafted, afford interpretability through intuitive explanations [15]. However, with deep learning models, interpretability is often sacrificed for performance gains achieved through abstraction (more layers) and integration (end-to-end training). Notably, recent advancements in Convolutional Neural Networks (CNNs) demonstrate state-of-the-art performance across diverse tasks but present challenges in interpretability due to their complexity [16, 17, 18]. Consequently, the field of deep learning is increasingly exploring the delicate balance between interpretability and accuracy, aiming to reconcile the benefits of sophisticated models with the need for transparency and comprehension. This balance is crucial not only for fostering trust in AI systems but also for facilitating the practical integration of these models into diverse sectors.

Counterfactual-based methods for generating visual explanations [19, 20] have provided valuable insights into the decision-making process of CNNs by identifying the most important features for classification tasks. However, these methods are often computationally expensive and sensitive to slight changes in input data. Visual explanation techniques, such as Class Activation Mapping (CAM) [21] and its extensions like Grad-CAM [22], Grad-CAM++ [23], and Layer-CAM [24], have become widely used for visualizing CNNs' decision-making processes. Each method offers a different approach to feature visualization, but they often fall short in certain applications [25], particularly when dealing with complex, high-dimensional data where interpretability is critical.

Several factors contribute to these limitations. Many techniques, excluding Layer-CAM, rely solely on the final convolutional layer to produce class activation maps. This approach may overlook valuable information from earlier layers that play a role in a model's perception. Moreover, the spatial resolution of the final layer's output is often low, which results in coarse activation maps that miss finer details. In contrast, shallow layers capture higher-resolution features, making them better suited for intricate detail detection.

Even when incorporating multiple layers, it is essential to recognize that not all layers hold the same significance within the model. While many CAM methods, including Grad-CAM, offer the flexibility to use layers beyond the final convolutional layer, they do not provide guidance on which layers are most informative for visualization. Furthermore, these methods do not provide insights into whether a specific layer is crucial for interpretation. This leaves layer selection to trial and error, adding complexity and inconsistency to the interpretation process [26].

For CNNs employing Global Average Pooling (GAP), prior CAM approaches represent the final classification score $Y^C$ for a class $C$ as a linear combination of the global average pooled final layer feature map $A^k$:

$$Y^C = \sum_k w_k^c \sum_{i,j} A_{ij}^k \tag{1}$$

where $w_k^c$ is the weight for each feature map $A^k$. However, this approach overlooks an essential aspect of CNNs: the presence of bias terms in fully connected (fc) layers. CNN models inherently rely on both weights and biases, with each contributing to the overall structure and accuracy of the network's predictions. Simply condensing feature maps into a weighted linear combination and neglecting the bias term can potentially limit the overall accuracy of these techniques, as they may miss subtle but significant biases present in the model. The presence of different bias terms for different channels can significantly impact the final class activation map for a layer, underscoring the importance of accounting for bias terms in CNN classifiers.

To capture this complexity more accurately, the final classification score $Y^C$ can be expressed as:

$$Y^C = \sum_k w_k^c \left( \sum_{i,j} A_{ij}^k \right) + b_k^c \tag{2}$$

where $b_k^c$ represents the bias term associated with each channel in the fc layer. Including this bias term refines the generated class activation maps, as it accounts for the network's full architecture, improving interpretability by incorporating both weights and biases—a fundamental part of CNN classifiers.

To address these limitations, this paper introduces a novel method called Integrative CAM (I-CAM), designed to enhance interpretability by leveraging key layers throughout the model for a comprehensive view of feature activations. I-CAM introduces a new layer score function that assesses each layer's relevance, removing ambiguity in layer selection. Additionally, I-CAM incorporates bias terms for each feature map $A^k$ to refine the classification score, as outlined in Equation 2, ensuring a more accurate reflection of the model's decision-making process.

Beyond these innovations, I-CAM improves generalizability through a simplified adaptation of the alpha term from Grad-CAM++. This modification replaces the complex $n^{th}$ order partial derivatives with a simpler expression involving the $n^{th}$ power of the first-order derivative, broadening applicability to any smooth function beyond exponentials. This enhancement not only streamlines I-CAM's implementation but also expands its utility across diverse functions.

## 2. Integrative CAM (I-CAM)

I-CAM is an innovative gradient-based Class Activation Mapping (CAM) method that redefines core principles in deep learning interpretability. It introduces an automated, dynamically weighted layer selection mechanism, moving beyond the conventional manual selection approach. By incorporating a bias term in saliency map computations and presenting a generalized alpha term from Grad-CAM++, I-CAM offers a more robust framework for representing feature importance in model decision-making [27]. This approach sets a new benchmark in AI interpretability, enhancing the understanding of model behavior and promoting the credibility of AI systems.

### 2.1. Perturbation-based layer importance score

The Integrative Class Activation Mapping (I-CAM) method employs a unique strategy to evaluate the relative importance of CNN layers. Each layer receives an importance score based on an analysis of the activation patterns and gradient information. However, it is essential to note that the determination of this layer-specific score cannot be solely reliant on a single input image and the associated output class probability. To provide a reliable layer scoring, I-CAM evaluates $n$ perturbed variations of the original input image $I$, yielding a set of $n + 1$ inputs. This multi-input framework enhances robustness by incorporating diverse perspectives from each perturbation, allowing a more nuanced understanding of feature activations [28, 29].

The importance score $S_l$ for layer $l$ is calculated as:

$$S_l = \sum_{i=1}^{n} w_i \sqrt{\sum \left( \left\| I \cdot \frac{\partial O}{\partial I} \right\|_F - \left\| \mathcal{G}_{i,l} \right\|_F \right)^2} \tag{3}$$

Here, $w_i$ is the weight for each perturbed input $I'_i$, $\frac{\partial O}{\partial I}$ represents the gradient of the output $O$ concerning the original input image $I$, $\|\cdot\|_F$ denotes Frobenius norm [30] along the channel dimension and $\mathcal{G}_{i,l}$ is the gradient-weighted activation for $I'_i$ at layer $l$. If $A$ and $\frac{\partial O}{\partial A}$ are the activation and gradient, respectively, of layer $l$, then $\mathcal{G}_l = relu\left(A \odot \frac{\partial O}{\partial A}\right)$.

### 2.1.1. Image perturbations

To generate the $n$ perturbed variations of the original image $I$, we apply a two-step perturbation process. First, random Gaussian noise $N_i$ is added to the image, where $N_i \sim \mathcal{N}(0,1)$ and scaled by a factor $\alpha$. This introduces variability into the image. Then, a random pixel mask $M_i$ is applied, which is sampled from a Bernoulli distribution $M_i \sim Bernoulli(1 - \alpha)$. This mask removes a proportion $1 - \alpha$ of the pixels uniformly across all channels.

The perturbed image $I'_i$ ($i = 1, 2, \cdots, n$) from the original image $I$ is given by:

$$I'_i = (I + \alpha N_i) \odot M_i \tag{4}$$

where $\odot$ denotes element-wise multiplication. The parameter $\alpha$ controls the intensity of the noise addition and pixel masking, making the perturbation process highly flexible. This dual operation, combining noise with random masking, ensures spatial consistency while allowing for meaningful perturbations of the image. Figure 1 illustrates eight such perturbations applied to an ImageNet [31] validation image with $\alpha = 0.4$.

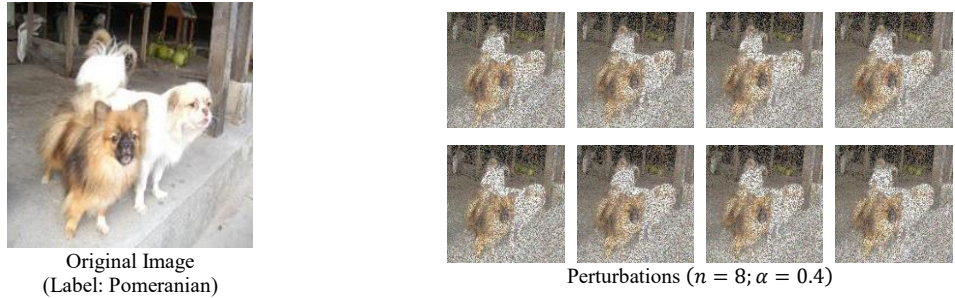

Original Image
(Label: Pomeranian)     Perturbations ($n = 8; \alpha = 0.4$)

Fig. 1: Perturbations on an image using random noise and pixel-masking

### 2.1.2. Perturbation weights

The generation and utilization of perturbations derived from an original image $I$ is crucial in the evaluation and robustness assessment of CNN classifiers. To acknowledge the varying importance and interactions of different perturbed images with the model, as not all perturbations will yield equally valuable insights, a more sophisticated and academically sound method is to employ a weighted mean.

The proposed methodology involves assigning a weighting factor $w_i$ to each perturbed input image $I'_i$, which is calculated based on two key considerations. Firstly, the SVIM metric between the original image $I$ and the perturbed image $I'_i$ captures the extent of the perturbation. Secondly, the similarity between the output class probabilities $O$ of the model for the original image $I$ and the output class probabilities $O'_i$ for the perturbed image $I'_i$, as quantified by the MDS metric, prioritizes perturbations that do not significantly alter the model's output.

The final weighting factor $w_i$ is calculated as the geometric mean of the SVIM and MDS values and ensures that the weighting factors are on a comparable scale, allowing for a meaningful aggregation of the perturbations. Thus, the perturbation weight $w_i$ is calculated as follows:

$$w_i = \sqrt{SVIM(I, I'_i) * MDS(O, O'_i)} \quad (5)$$

### 2.1.3. Structural Variability Index Metric (SVIM)

The SVIM (Structural Variability Index Metric) function is a Gaussian-based metric derived from the well-established SSIM (Structural Similarity Index Metric) [32]. While the SSIM metric theoretically ranges from -1 to 1, with 1 indicating a perfect match, in practical cases, it is bound to the range [0, 1]. The SVIM metric focuses on quantifying the variability of the structures between two input images, providing a measure of how much the structural components of the perturbed image differ from those of the original image. A value of 0 indicates that the perturbed image is either identical to or completely dissimilar from the original, while a value of 1 signifies that the perturbed image presents a meaningful variation, maintaining structural similarity but with some degree of difference. The SVIM function is defined as:

$$SVIM(x, y) = e^{-\frac{(SSIM(x,y) - 0.5)^2}{2\sigma^2}} \quad (6)$$

where SSIM is computed as:

$$SSIM(x, y) = \frac{(2\mu_x \mu_y + c_1)(2\sigma_{xy} + c_2)}{(\mu_x^2 + \mu_y^2 + c_1)(\sigma_x^2 + \sigma_y^2 + c_2)} \quad (7)$$

Here, $\mu_x$, $\mu_y$ are the means of the images, $\sigma_x^2$, $\sigma_y^2$ are the variances, $\sigma_{xy}$ is the covariance, and $c_1$, $c_2$ are small positive constants used for numerical stability.

By capturing both structural similarity and variability, the SVIM metric provides a more nuanced measure of image differences compared to traditional difference-based metrics. This makes it highly effective in assessing how meaningful a perturbation is, where a high SVIM score indicates a good perturbation that introduces useful variation without straying too far from the original structure.

### 2.1.4. Mean Difference Similarity (MDS)

The MDS (Mean Difference Similarity) function introduces an innovative approach for quantifying similarity between two probability distributions, $X$ and $Y$, and diverges from traditional metrics such as Jensen-Shannon Divergence (JSD) [33]. While JSD relies on computing the average Kullback-Leibler (KL) divergence [34] between input distributions and an intermediary distribution, the MDS function bypasses this by employing a direct comparison between the distributions. At its core, MDS calculates similarity based on the Mean Difference Divergence (MDD). a weighted average of element-wise ratios between $X$ and $Y$.

The MDD function is formulated by taking the weighted mean of the differences between corresponding elements $X_k$ and $Y_k$, with each weight $D_k = \frac{X_k - Y_k}{2}$. This results in a weighted divergence score that is subtracted from 1 to yield the MDS similarity score. Mathematically, MDD and MDS are defined as follows:

$$MDD(X,Y) = \sum_{k=1}^{C} D_k \, log\left(\frac{X_k}{Y_k}\right) \tag{8}$$

$$MDS(X,Y) = 1 - MDD(X,Y) \tag{9}$$

where $X$ and $Y$ are the input probability distributions, $C$ is the number of elements in each distribution, and $X_k$ and $Y_k$ are the $k^{th}$ elements of each distribution.

By comparing distribution elements directly, the MDS function avoids dependence on a mid-point distribution, thus simplifying the calculation and eliminating assumptions about intermediary forms that may impact the resulting similarity measure. Alternatively, $MDD(X,Y)$ can also be conceptualized as a form of symmetric Jensen-Shannon Divergence, computed as the average KL divergence in both directions:

$$MDD(X,Y) = \frac{KL(X,Y) + KL(Y,X)}{2} \tag{10}$$

The symmetric property of MDS, where $MDS(X,Y) = MDS(Y,X)$, aligns with the intuitive concept of similarity, as the similarity degree between two distributions should be consistent regardless of order. As a whole, the MDS function offers a compelling alternative to traditional metrics like JSD, providing a direct, interpretable measure of similarity between probability distributions that is applicable across a wide range of research domains.

## 2.2. Redundant layer filtering and layer weightage

Consider a CNN model with $L$ layers. To prioritize meaningful feature extraction across these layers, we assign an importance score $S_l$ to each layer $l$ ($l \in \{1, 2, ..., L\}$), resulting in a list of layer scores $S = \{S_1, S_2, ..., S_L\}$. Some of these layers may contribute minimal information and can be considered redundant for the image at hand with respect to the model's prediction [35, 36, 37]. To address this, we rank the layers in descending order of their importance scores, forming a new ordered list $S' = \{S'_1, S'_2, ..., S'_L\}$ such that $S'_1 \geq S'_2 \geq \cdots \geq S'_L$. We then compute the cumulative sum of these importance scores as follows:

$$C_k = \sum_{l=1}^{k} S'_l \tag{11}$$

where $C_k$ represents the cumulative importance score up to the $k^{th}$ layer. To ensure that only the most informative layers are retained, we set a threshold $T$, representing the percentage of the overall importance score to be preserved. In this study, we use a threshold of 0.95 (or 95%), which captures the majority of critical features for effective feature map fusion for model perception visualization. We determine the index $k$ by satisfying the following conditions:

$$C_k \geq T * \sum_{l=1}^{L} S'_l \tag{12}$$

$$C_{k-1} < T * \sum_{l=1}^{L} S_l' \tag{13}$$

This means that the cumulative score up to the $k^{th}$ layer meets or exceeds the threshold, while the score up to the $(k-1)^{th}$ layer remains below it. Once we identify the top $k$ layers to retain, we allocate weights to these selected layers based on their relative importance scores. The weight $W_l$ for the $l^{th}$ layer is defined as follows:

$$W_l = \frac{S_l}{\sum_{j \in L'} S_j} \tag{14}$$

where $L'$ represents the subset of the $k$ selected layers from the original $L$. This process enables focused integration of the most informative layers, enhancing the effectiveness of feature fusion by emphasizing layers that significantly contribute to the model's interpretability and performance.

## 2.3. Gradient-based CAM approaches

Gradient-based CAM methods such as Grad-CAM and Grad-CAM++ help localize target objects in images by computing weighted sums of feature maps from the final convolutional layer, where weights represent each map's importance for the target class. In Grad-CAM, these weights, $w_k^c$, are determined by averaging gradients over spatial locations:

$$w_k^c = \frac{1}{N} \sum_{i,j} g_{ij}^{kc} \tag{15}$$

where $N$ is the number of locations, $Y^c$ is the model output, and $g_{ij}^{kc} = \frac{\partial Y^c}{\partial A_{ij}^k}$ is the gradient at location $(i,j)$. Grad-CAM++ enhances this by incorporating higher-order gradients, refining $w_k^c$ with spatially-dependent terms:

$$w_k^c = \sum_{i,j} \alpha_{ij}^{kc} \cdot relu(g_{ij}^{kc}) \tag{16}$$

Here, $\alpha_{ij}^{kc}$ is a factor derived from second- and third-order gradients:

$$\alpha_{ij}^{kc} = \frac{\frac{\partial^2 Y^c}{(\partial A_{ij}^k)^2}}{2\frac{\partial^2 Y^c}{(\partial A_{ij}^k)^2} + \sum_{i,j} A_{ij}^k \left\{\frac{\partial^3 Y^c}{(\partial A_{ij}^k)^3}\right\}} \tag{17}$$

Following the computation of the channel-wise weights $w_k^c$ for each feature map $A^k$, a rectified linear unit (ReLU) operation is employed to eliminate negative responses to get the resulting class activation map $M^c$, defined as:

$$M^c = relu\left(\sum_k w_k^c \cdot A_k\right) \tag{18}$$

While these methods can be applied to both shallow and deep layers, they use a global weighting scheme across spatial locations, which works well for deeper layers where feature maps are smaller and spatial variance is low. However, applying these global weights to shallow layers, where feature maps are larger and spatial variance is high, can lead to suboptimal results. Shallow layers capture more fine-grained and localized details, making it important to account for spatially varying contributions rather than a single global weight. LayerCAM mitigates this by applying localized, element-wise gradients as weights at each spatial location:

$$w_{ij}^{kc} = relu(g_{ij}^{kc}) \tag{19}$$

This element-wise weighting allows LayerCAM to emphasize important local features within each feature map, making it better suited for use on shallow layers that capture detailed patterns. However, LayerCAM's method of averaging CAMs from all layers to obtain the final map may dilute the contribution of layers with varying relevance, potentially limiting interpretability.

To address this limitation, our proposed Integrative CAM (I-CAM) method extends Grad-CAM++ by combining LayerCAM's element-wise approach with adaptive layer selection. I-CAM computes CAMs for the top 95% of layers based on their importance scores and weights each layer's CAM according to its specific relevance, avoiding a simple average across all layers. Additionally, I-CAM introduces a bias term, $b_k^c$, to capture more nuanced aspects of the model's behavior. The class activation map $L^c$ for a selected layer using the I-CAM method is defined as:

$$L^c = \sum_k w_{ij}^{kc} \cdot A_{ij}^k + b_k^c \tag{20}$$

The final I-CAM, $L_{I-CAM}^c$, aggregates these layer-specific CAMs, weighted by the importance $W_l$ of each layer $l$:

$$L_{I-CAM}^c = \sum_{l \in L'} W_l L_l^c \tag{21}$$

where $W_l$ represents the weight for layer $l$ among the selected layers $L'$. By integrating spatial importance with adaptive layer-based weighting, I-CAM provides a more comprehensive view of the model's decision-making process, enhancing interpretability across the CNN and improving the quality of class activation maps for more accurate visualization.

### 2.4. Alpha term simplification

In classification models, the penultimate score for class $c$, denoted as $S^c$, is a fundamental element that influences the final class prediction. The general form of a classification head, outlined in Eq. 22, may include linear or ReLU activations to compute $S^c$. The parameters $\omega_k^c$ and $\beta_k^c$ represent the activation weight and bias of the classifier, respectively, while $h(A^l)$ refers to the transformation applied to the activation $A^l$ from layer $l$ before passing through the classifier. If $l$ is the final layer, then $h(A^l) = A^l$.

$$S^c = g(A_{ij}^{kl}) = \sum_k \omega_k^c \sum_{i,j} h(A_{ij}^{kl}) + \beta_k^c \tag{22}$$

Gradients, such as $\frac{\partial S^c}{\partial A_{ij}^{kl}} = \omega_k^c \cdot h'(A_{ij}^{kl})$, can be computed with automatic differentiation tools like PyTorch. If $h(A_{ij}^{kl})$ is a linear function of $A_{ij}^{kl}$, potentially including ReLU activations, the derivative $h'(A_{ij}^{kl})$ is constant. Consequently, higher-order derivatives $\frac{\partial^2 S^c}{\partial (A_{ij}^{kl})^2}$ and $\frac{\partial^3 S^c}{\partial (A_{ij}^{kl})^3}$ will be zero.

Grad-CAM directly uses this $S^c$ value as the class score $Y^c$ without further transformation. However, in most cases, the predicted output $Y^c$ is not taken directly from $S^c$; instead, it undergoes a smoothing transformation, such as an exponential or softmax function, to ensure a stable and differentiable output. Grad-CAM++, for instance, applies an exponential transformation to $S^c$ to make $Y^c$ infinitely differentiable, which is beneficial for obtaining consistent higher-order derivatives necessary for easier implementation of their alpha expression. Expanding on this idea, we derive a general approach to express higher-order derivatives as functions of the first-order derivative for any smooth function $f$, allowing consistent application across interpretability methods that rely on multi-order derivatives. If $Y^c = f(S^c)$, is the final score after passing model output $S^c$ through a smooth function $f$, then:

$$\frac{\partial Y^c}{\partial A_{ij}^{kl}} = \frac{\partial Y^c}{\partial S^c} \cdot \frac{\partial S^c}{\partial A_{ij}^{kl}}$$

$$\frac{\partial Y^c}{\partial A_{ij}^{kl}} = f'(S^c) \cdot \frac{\partial S^c}{\partial A_{ij}^{kl}} \qquad (23)$$

$$\frac{\partial^2 Y^c}{\partial (A_{ij}^{kl})^2} = \frac{\partial}{\partial A_{ij}^{kl}} \left( \frac{\partial Y^c}{\partial A_{ij}^{kl}} \right) = \frac{\partial}{\partial A_{ij}^{kl}} \left( f'(S^c) \cdot \frac{\partial S^c}{\partial A_{ij}^{kl}} \right) = \frac{\partial S^c}{\partial A_{ij}^{kl}} \cdot \frac{\partial f'(S^c)}{\partial A_{ij}^{kl}}$$

$$= \frac{\partial S^c}{\partial A_{ij}^{kl}} \cdot \frac{\partial f'(S^c)}{\partial S^c} \cdot \frac{\partial S^c}{\partial A_{ij}^{kl}} = f''(S^c) \cdot \left( \frac{\partial S^c}{\partial A_{ij}^{kl}} \right)^2$$

$$\frac{\partial^3 Y^c}{\partial (A_{ij}^{kl})^3} = \frac{\partial}{\partial A_{ij}^{kl}} \left( \frac{\partial^2 Y^c}{\partial (A_{ij}^{kl})^2} \right) = \frac{\partial}{\partial A_{ij}^{kl}} \left( f''(S^c) \cdot \left( \frac{\partial S^c}{\partial A_{ij}^{kl}} \right)^2 \right) = \left( \frac{\partial S^c}{\partial A_{ij}^{kl}} \right)^2 \cdot \frac{\partial f''(S^c)}{\partial A_{ij}^{kl}}$$

$$= \left( \frac{\partial S^c}{\partial A_{ij}^{kl}} \right)^2 \cdot \frac{\partial f''(S^c)}{\partial S^c} \cdot \frac{\partial S^c}{\partial A_{ij}^{kl}} = f'''(S^c) \cdot \left( \frac{\partial S^c}{\partial A_{ij}^{kl}} \right)^3$$

Thus ultimately, for $Y^c = f(S^c)$,

$$\frac{\partial^n Y^c}{\partial (A_{ij}^{kl})^n} = f^n(S^c) \cdot \left( \frac{\partial S^c}{\partial A_{ij}^{kl}} \right)^n \qquad (24)$$

Hence, Eq. 16 is reduced to the form:

$$\alpha_{ij}^{kc} = \frac{f''(S^c) \cdot \left( \frac{\partial S^c}{\partial A_{ij}^{k}} \right)^2}{2f''(S^c) \cdot \left( \frac{\partial S^c}{\partial A_{ij}^{k}} \right)^2 + \sum_{i,j} A_{ij}^{k} \left\{ f'''(S^c) \cdot \left( \frac{\partial S^c}{\partial A_{ij}^{k}} \right)^3 \right\}} \qquad (25)$$

This simplification is only feasible under the assumption that the partial derivatives of the objective function $S^c$ for the layer activation $A_{ij}^{kl}$ exhibit a specific behavior. Specifically, the assumption is that $\frac{\partial^n S^c}{\partial (A_{ij}^{kl})^n} = 0$ for $n > 1$, meaning that the first-order partial derivative $\frac{\partial S^c}{\partial A_{ij}^{kl}}$ is a constant and not a function of the $A_{ij}^{kl}$. This holds when $h(A_{ij}^{kl})$ is a linear function. When this assumption is held, simplification can be conducted. However, if $h(A_{ij}^{kl})$ is a non-linear function, it can still be approximated by a piecewise linear function $F(A_{ij}^{kl})$, such that $h(A_{ij}^{kl}) = F(A_{ij}^{kl}) + \varepsilon$ (where $\varepsilon$ represents the approximation error). The simplification then becomes valid upon neglecting the approximation error. Fortunately, many common non-linear activation functions, such as tanh and sigmoid, can already be expressed in piecewise linear forms hard-tanh and hard-sigmoid, respectively. This allows for the valid application of the simplification to the alpha term in Eq. 25, as the required assumptions are met.

For some other non-linear functions that involve more complicated behavior, like $GeLU(x)$ we can still break it to a piece-wise linear form as any function can be approximated into piecewise linear functions. Thus, any non-linear function can be converted to a piece-wise linear form and the simplification of the alpha term becomes valid if approximation error ($\varepsilon$) is low.

## 2.5. Weight and bias calculation

To determine the gradient-based weights for a layer, we integrate methodologies from Grad-CAM++ and LayerCAM. Starting with Eqs. 16 and 19, we derive the following equation for the weight term:

$$w_{ij}^{kc} = tanh(\alpha_{ij}^{kc}) \cdot relu(g_{ij}^{kc}) = tanh(\alpha_{ij}^{kc}) \cdot relu\left(\frac{\partial Y^c}{\partial A_{ij}^{kl}}\right)$$

Substituting value of $\frac{\partial Y^c}{\partial A_{ij}^{kl}}$ from Eq. 23,

$$w_{ij}^{kc} = tanh(\alpha_{ij}^{kc}) \cdot relu\left(f'(S^c) \cdot \frac{\partial S^c}{\partial A_{ij}^k}\right) \quad (26)$$

This allows us to compute the weight term for an activation map using the alpha value from Eq. 25, along with the gradients of the activation and the smooth function.

For the bias term, considering that the model output $S^c$ depends on a linear combination of activations from the $l^{th}$ layer, we can express it as:

$$S^c = \sum_{i,j} w_{ij}^{kc} \sum_{i,j} A_{ij}^k + b_k^c$$

We present two alternative approaches for defining the bias term: as a uniform value applied across an entire feature channel or as an element-wise term specific to individual spatial locations. These approaches yield the following expressions for the bias term:

$$b_k^c = S^c - \sum_{i,j} w_{ij}^{kc} \sum_{i,j} A_{ij}^k \quad (27)$$

$$b_k^c = S^c - w_{ij}^{kc} \sum_{i,j} A_{ij}^k \quad (28)$$

Eq. 27 represents a uniform bias factor applied to an entire channel, functioning as a generalized bias term, which is more in line with how actual bias works in a model. In contrast, Eq. 28 treats the bias as a spatially varying feature map, assigning distinct bias values to specific locations within the map. Since this transformation approach reduces two-dimensional feature maps to a single dimension via Global Average Pooling (GAP), the applicability of a uniform bias across all pixel locations in a channel requires validation. Drawing from LayerCAM's approach, which utilizes element-wise gradients for enhanced spatial detail, we explore these two bias formulations to evaluate their impact on localized feature importance. By experimentally comparing these formulations, we aim to identify the optimal bias treatment for improving feature integration and interpretability in class activation mapping.

## 2.6. SoftMax function as a smooth function

In our methodology, we draw inspiration from Grad-CAM++, which utilizes the exponential function for its smoothing properties, by employing the softmax function. The softmax function is a well-established method for deriving final class probabilities in classification tasks. Using the softmax function, the final class score $Y^c$ is defined as follows:

$$Y^c = f(S^c) = \frac{e^{S^c}}{\sum_i e^{S^i}} \quad (29)$$

Here, the index $i$ runs over all the output classes, and $S^i$ is the score associated with output class $i$ in the penultimate layer. The derivatives of the softmax function for $S^c$ are given by:

$$\frac{\partial Y^c}{\partial S^c} = f'(S^c) = Y^c(1 - Y^c)$$

$$\frac{\partial^2 Y^c}{\partial (S^c)^2} = f''(S^c) = Y^c(1 - 3Y^c + 2(Y^c)^2)$$

$$\frac{\partial^3 Y^c}{\partial (S^c)^3} = f'''(S^c) = Y^c(1 - 7Y^c + 12(Y^c)^2 - 6(Y^c)^3)$$

By incorporating the SoftMax function, the alpha term and the weight term are expressed as:

$$\alpha_{ij}^{kc} = \frac{Y(1 - 3Y + 2Y^2) \cdot g^2}{2Y(1 - 3Y + 2Y^2) \cdot g^2 + \sum_{i,j} A_{ij}^k \{Y(1 - 7Y + 12Y^2 - 6Y^3) \cdot g^3\}} \quad (30)$$

$$w_{ij}^{kc} = tanh(\alpha_{ij}^{kc}) \cdot relu(Y(1 - Y) \cdot g) \quad (31)$$

Here $g = \frac{\partial S^c}{\partial A_{ij}^k}$ and $Y = Y^c$. These formulations leverage the smoothing properties of the softmax function to enhance the interpretability and effectiveness of our visualization method.

## 3. Experiments and Results

To rigorously evaluate the performance and advantages of the proposed Integrative CAM (I-CAM) method, we conducted a series of experiments on the well-established ImageNet dataset, using the ResNet-50 architecture. This section provides an overview of the experimental setup, including dataset details, model configurations, and preprocessing procedures. By benchmarking I-CAM against other similar gradient-based techniques, our goal is to demonstrate its robustness and accuracy in the context of class activation mapping (CAM) and explainable AI (XAI).

Both quantitative and qualitative assessments were carried out to comprehensively evaluate the effectiveness of I-CAM. Quantitative metrics such as Intersection over Union (IoU) were used to measure the localization accuracy of generated heatmaps, while saliency scores were computed to assess the method's ability to highlight relevant image regions. Qualitative analysis included visual comparisons to evaluate the interpretability and fidelity of the heatmaps produced by I-CAM. Furthermore, we conducted ablation studies to examine the contribution of each component of the method, alongside an analysis of computational efficiency to assess scalability and practical applicability in real-world scenarios.

The ImageNet dataset, known for its diversity and complexity, was selected to ensure a robust evaluation across various contexts, offering insights into I-CAM's interpretability and generalization capabilities. For consistency, experiments were performed using a pre-trained ResNet-50 model, with specific configurations adapted based on the experimental objectives.

Through these extensive experiments, we provide strong evidence for the superior interpretability and visualization quality of I-CAM, emphasizing its robustness in both qualitative and quantitative evaluations. To ensure reliability, each method, including I-CAM, was evaluated over multiple runs (10 iterations) to account for variability. The following sections describe the experimental details, including datasets, implementation specifics, quantitative and qualitative results, and ablation studies, offering a comprehensive and transparent view of I-CAM's strengths and practical applications.

### 3.1. Qualitative Analysis

In this section, we analyze the quality of the proposed I-CAM method with respect to its sensitivity to bias, a comparison with existing gradient-based CAM techniques, and the impact of layer weight selection.

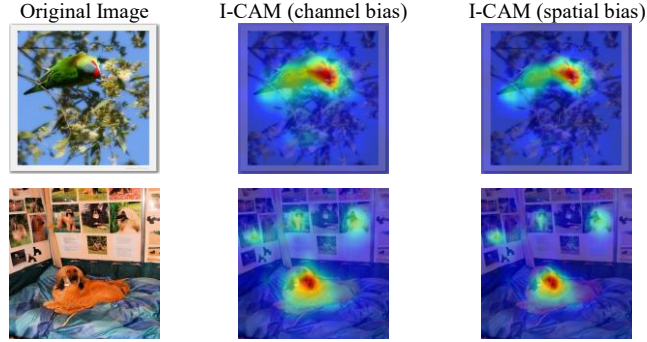

Fig. 2: Examples of I-CAM outputs using channel bias and spatial bias overlayed on their corresponding images

### 3.1.1. Spatial vs Channel Bias Impact

Fig. 2 presents examples of I-CAM outputs with both channel and spatial biases superimposed on the corresponding images. The I-CAM method supports two bias options: channel-wise bias (Eq. 27) and spatial bias (Eq. 28). This section evaluates the performance of I-CAM using these two bias configurations, comparing their respective outcomes. The evaluation was conducted on images from the ImageNet validation set, utilizing a ResNet-50 model pre-trained on the ImageNet training dataset.

For the comparative analysis, a blind survey was conducted with a subset of 50 images chosen from the 50,000 ImageNet validation images. These images were selected from five classes, with each class contributing 10 images having the highest confidence scores. The classes (chickadee, lorikeet, spoonbill, Afghan hound, and carousel) were selected based on the highest F1-scores.

The survey involved 10 participants from diverse professional backgrounds, with varying levels of familiarity with CNNs and AI. To minimize bias, the order of presentation for the outputs was randomized, alternating between spatial and channel bias. Each method was scored based on the participants' evaluations.

To ensure fairness, the scores for each image were normalized, with the highest possible score set to 1.0. The total normalized scores resulted in a maximum score of 50. In this analysis, channel bias achieved a score of 33.8 (67.6% of the total), while spatial bias scored 16.2 (32.4% of the total). These results indicate that the channel bias offers a more significant enhancement to the I-CAM method compared to the spatial bias.

### 3.1.2. Comparison with other methods

To assess the effectiveness and interpretability of the proposed I-CAM method, a detailed qualitative analysis was performed. This subsection provides visual comparisons between I-CAM and existing gradient-based techniques, including Grad-CAM, Grad-CAM++, and LayerCAM. Similar to the evaluation of bias impact, the analysis incorporates a visual survey to compare the performance of these methods. Fig. 3 illustrates the outputs of all algorithms superimposed on the corresponding images. For the I-CAM method, the channel bias formulation was chosen for comparison, as it demonstrated superior performance over the spatial bias option.

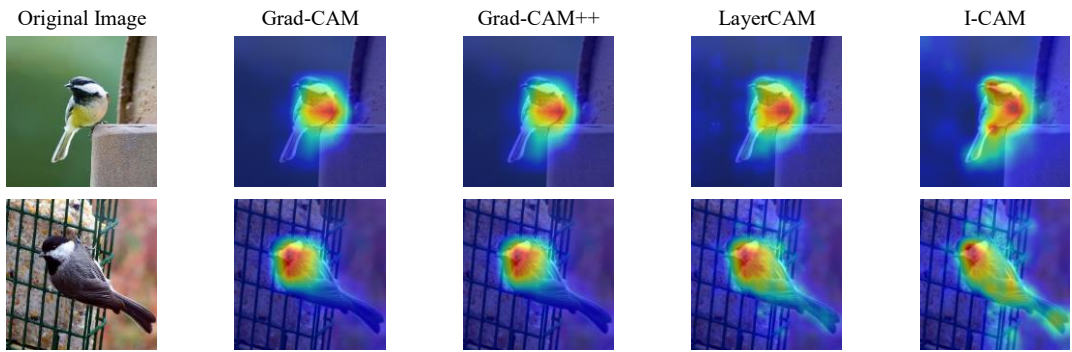

Fig. 3: Examples of outputs of all methods overlayed on their corresponding images

A blind survey, involving a different set of 10 participants, was conducted using the same collection of 50 images as in the previous analysis. To minimize bias, the outputs from the four methods were shown in a randomized order. Participants were tasked with selecting the image that best represented the target object in each case. Each method received a score based on participant responses.

The scores for each image were normalized and aggregated to achieve a total possible score of 50 per method. In this analysis, I-CAM achieved a score of 24.4, representing 48.8% of the total. In comparison, LayerCAM, Grad-CAM++, and Grad-CAM received scores of 11.37, 7.76, and 6.47, corresponding to 22.74%, 15.52%, and 12.94%, respectively. These findings suggest that I-CAM provides superior visualizations for understanding the model's perception, offering clearer localization of key features compared to other gradient-based methods. This enhanced interpretability fosters greater confidence in selecting appropriate models for feature extraction tasks.

### 3.1.3. Model confidence analysis

Visual examples are presented to demonstrate the performance of the proposed method in identifying relevant regions within images and generating corresponding heatmaps. For each correct prediction, a saliency score ($\mathcal{S}$) is calculated, indicating the extent to which the heatmap covers the actual object in the image.

$$\mathcal{S} = \frac{\sum_{i,j} L_{I-CAM}^c \circ L_{Bbox}^c}{\sum_{i,j} L_{I-CAM}^c} \quad (32)$$

To thoroughly investigate the model's functionality, we designed our experiments around four distinct scenarios, each providing unique insights into the model's behavior. These scenarios were specifically selected to capture a broad range of model responses, allowing for a comprehensive evaluation of its performance under various levels of prediction confidence and accuracy. This structured approach facilitates a detailed analysis and interpretation of the model's decision-making process through the proposed visualization method, providing valuable insights into its operational dynamics.

**Case 1:**   **High Confidence, Correct Prediction**

In this scenario, the model correctly predicts the target class with high confidence, assigning a significantly higher probability to the predicted class than to any other. This outcome highlights the model's reliability and precision when it demonstrates strong decision-making confidence.

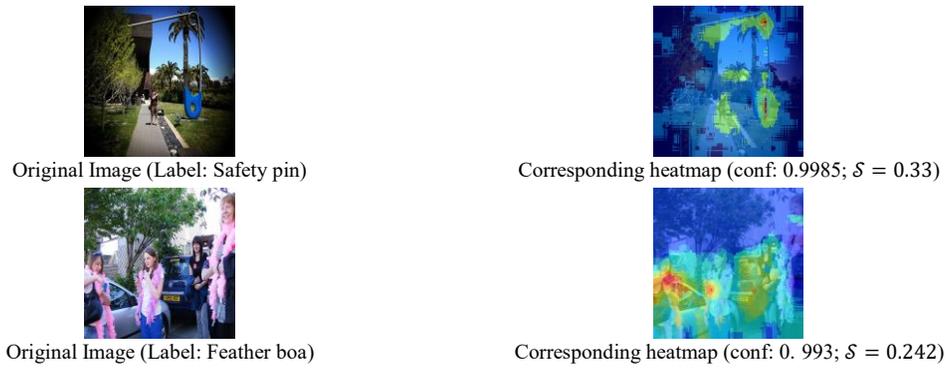

Original Image (Label: Safety pin)     Corresponding heatmap (conf: 0.9985; $S = 0.33$)

Original Image (Label: Feather boa)     Corresponding heatmap (conf: 0. 993; $S = 0.242$)

Fig. 4: Case 1 scenarios where model gives correct prediction with high confidence

Fig. 4 presents two examples: one labeled as "safety pin" and the other as "feather boa," along with their corresponding heatmaps. In the "safety pin" example, the heatmap emphasizes areas of higher activation around the head and coil sections of the pin, demonstrating the model's focus on these key features. In contrast, the "feather boa" image, which depicts multiple individuals wearing feather boas, shows the highest activation on the feather boas worn by two girls on the left. However, the model partially overlooks the girl on the far right, although her feather boa displays a slight activation. Additionally, a high activation is observed on the girl in the blue coat, which is an error, but the pattern follows similar behavior seen in the other correct detections of feather boas.

A key observation in this case is that images with correct predictions and high class-confidence (above 0.98) tend to exhibit a patchy appearance in their heatmaps. This effect becomes more pronounced as confidence increases, offering additional insights into the model's interpretability and decision-making. However, it is important to note that this patchiness may be influenced by artifacts resulting from heatmap resizing and normalization processes.

Specifically, the heatmap from the final convolutional layer, originally sized at (7,7), is resized to (224,224) to match the dimensions of the input image. The resizing process can introduce interpolation artifacts due to the significant enlargement of the relatively small heatmap. Furthermore, the normalization applied to the heatmap may also contribute to the observed patchy appearance. These combined factors—resizing and normalization—could distort the heatmap, leading to irregularities. While this patchiness provides valuable insights into the model's inner workings, it is essential to consider these potential artifacts when interpreting the heatmap behavior.

**Case 2:**     **Low Confidence, Correct Prediction**

The second scenario examines cases where the model correctly predicts the class, but with low confidence. In these cases, the probability assigned to the predicted class is only marginally higher than those assigned to other classes. This situation highlights the model's uncertainty and provides valuable insights into its decision-making process when faced with ambiguous or challenging inputs, even though the final classification is correct.

Fig. 5 shows two example images—one labeled "Egyptian cat" and the other "snail"—along with their corresponding heatmaps. The heatmaps, generated using the I-CAM method, demonstrate that while the model successfully identifies and localizes the object within the image, its confidence remains low. The I-CAM method, by providing detailed feature extraction visualizations, offers a comprehensive view of the model's internal processing. This enhanced transparency is particularly useful for understanding and interpreting the model's decision-making process.

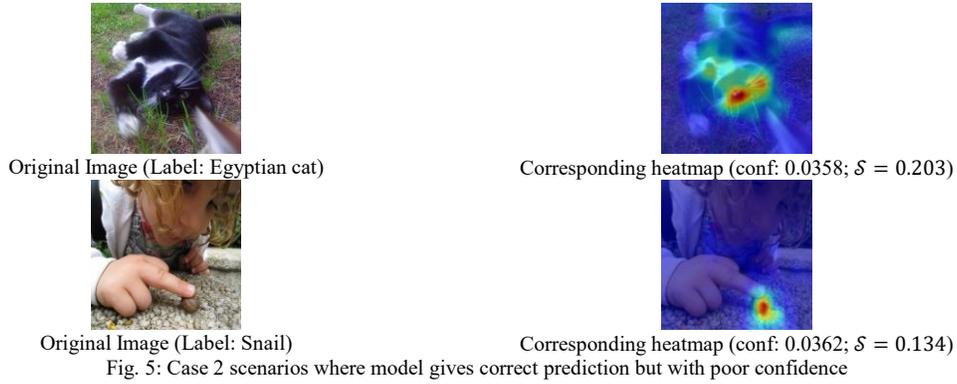

Original Image (Label: Egyptian cat)     Corresponding heatmap (conf: 0.0358; $\mathcal{S} = 0.203$)

Original Image (Label: Snail)     Corresponding heatmap (conf: 0.0362; $\mathcal{S} = 0.134$)

Fig. 5: Case 2 scenarios where model gives correct prediction but with poor confidence

During training, all layers of the model are updated based on the output, allowing the intermediate layers to learn and extract features relevant to the final decision. However, only the output of the final convolutional layer is passed to the fully connected layer for classification. This output represents a coarse and approximate summary of the model's behavior, which may contribute to the low confidence observed in the prediction. This also suggests that while the model's feature extraction aligns with human perception (i.e., "what you see is what you get"), the extracted features may not be sufficiently distinctive to enable confident class differentiation.

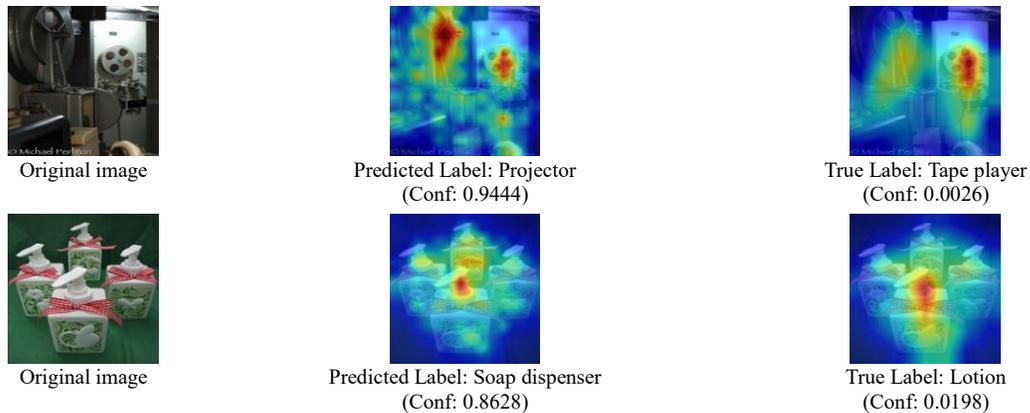

Original image     Predicted Label: Projector (Conf: 0.9444)     True Label: Tape player (Conf: 0.0026)

Original image     Predicted Label: Soap dispenser (Conf: 0.8628)     True Label: Lotion (Conf: 0.0198)

Figure 6: Case 3 scenarios where model gives incorrect prediction with high confidence

**Case 3:**     **High Confidence, Incorrect Prediction**

In this section, we analyze cases where the model confidently misclassifies an image, assigning a high probability to an incorrect class label. Understanding these high-confidence errors is essential to diagnosing the model's limitations and identifying potential biases.

In Fig. 6, two such examples are presented: one where an image labeled "Tape player" is incorrectly classified as "Projector," and another where an image labeled "Lotion" is misclassified as "Soap dispenser." To explore the model's reasoning, we use the Integrated Class Activation Mapping (I-CAM) method to generate heatmaps for both the predicted and true classes. These heatmaps reveal the regions of the image that the model relies on for classification, highlighting areas of similarity in features associated with both classes. For instance, in the "Tape player" image, the model focuses on components that resemble features of both a tape player and a projector, suggesting possible visual overlap. Similarly, for the "Lotion" example, the model misinterprets the handle-like structures of the lotion bottle as indicative of a soap dispenser.

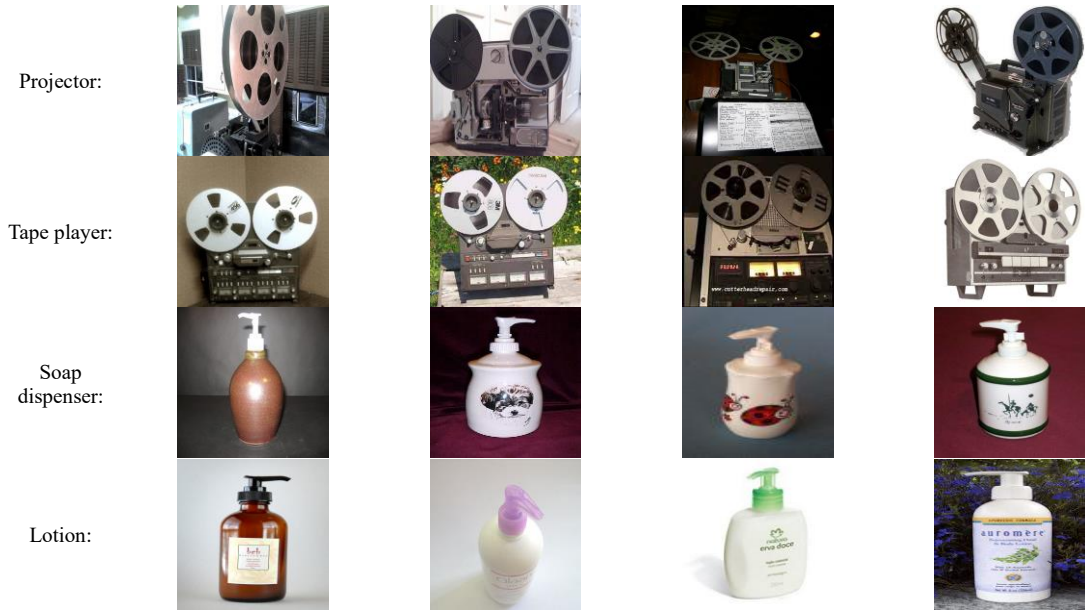

Figure 7: Images from ImageNet training dataset for understanding model's behavior for respective classes in Case 3

However, while the heatmaps reveal overlapping feature attention, this alone does not confirm that similar images in the training set could have contributed to these confusions. To investigate this further, we conducted an example-based comparison by reviewing representative training images for both true and predicted classes. Fig. 7 illustrates visually similar examples of tape players and projectors, as well as lotion bottles and soap dispensers, from the training data. These comparisons confirm that the training set does indeed contain similar images for these pairs of classes, which likely contributes to the model's misclassifications by reinforcing feature overlaps.

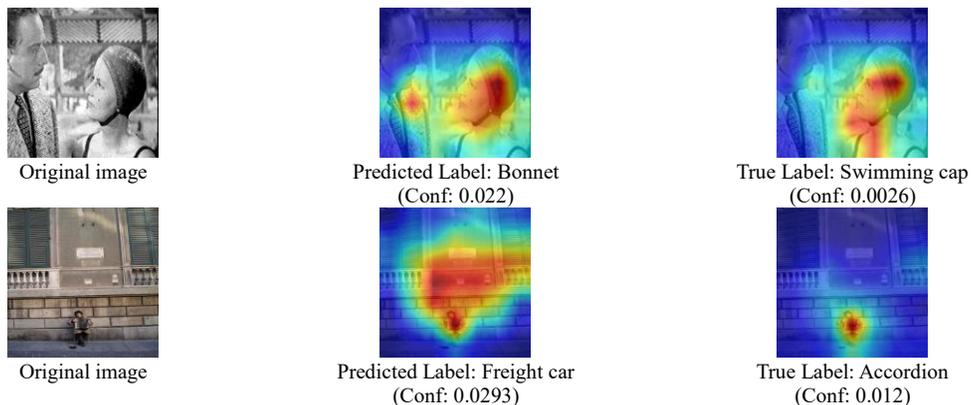

Figure 8: Case 4 scenarios where model performs the worst with incorrect prediction and poor confidence

This combined approach—first using heatmaps to reveal the model's feature focus, and then validating with example-based comparisons—provides a clearer understanding of how visual similarities in the training data may cause data uncertainty [38] and lead to high-confidence errors. Such insights can inform refinements to the model, ultimately enhancing its ability to distinguish between visually similar classes and reducing high-confidence misclassifications.

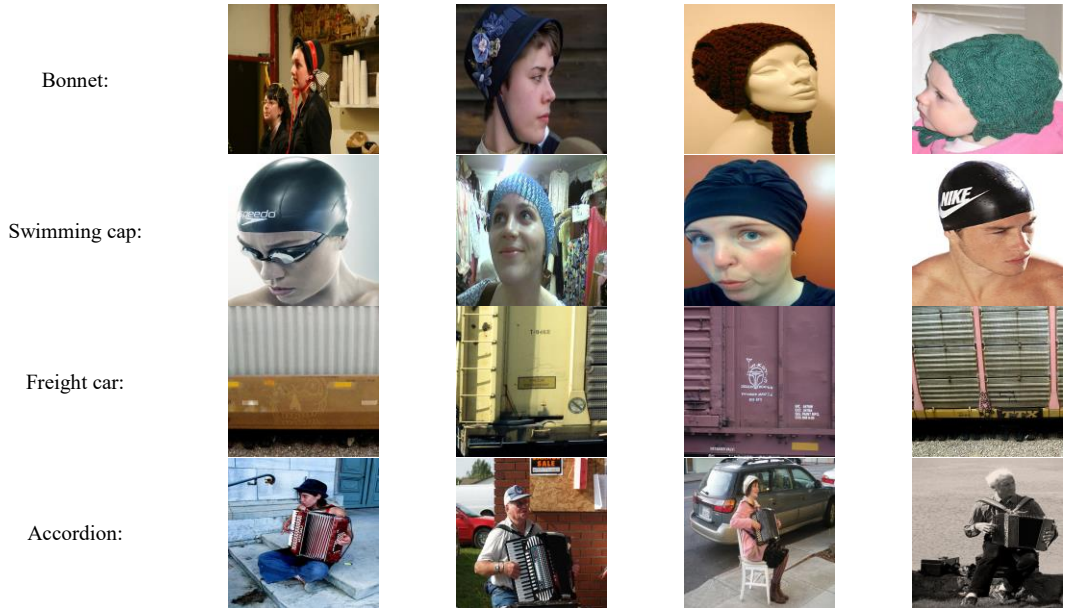

Figure 9: Images from ImageNet training dataset for understanding model's behavior for respective classes in Case 4

**Case 4:        Low Confidence, Incorrect Prediction**

This section examines cases where the model predicts an incorrect class with low confidence, assigning a probability to the predicted class that is comparable to other classes but still higher than the true class. Analyzing such cases is essential for understanding how the model interprets challenging inputs and where it struggles to distinguish between visually similar classes, resulting in misclassification.

In Fig. 8, two examples illustrate this behavior: an image labeled "Swimming cap" incorrectly predicted as "Bonnet," and another labeled "Accordion" misclassified as "Freight car." The I-CAM heatmaps are used here to visualize the parts of each image that the model considers important for both the true and predicted classes. Following this, example-based comparisons with similar training images (Fig. 9) provide further context, clarifying yet again how data uncertainty caused by visual similarities between classes in the training data may contribute to the model's confusion.

For the "Swimming cap" image, the heatmaps show that the model primarily focuses on the swimmer, with only a slight emphasis near the cap itself, making it insufficient for precise classification. The heatmap also shows some attention to the attire of the person on the left of the swimmer. It is through example-based comparisons with training images that we can further understand this misclassification. The training set for swimming caps often lacks the string detail seen in the test image, a feature more common in bonnets, leading to the model's confusion. Similarly, the texture of the attire of the person on the left resembles bonnet textures in the training data.

In the "Accordion" example, the model's interpretation similarly appears misguided but explainable. The heatmap for the true class "Accordion" shows the model focuses on relevant parts of the accordion, as well as on the person playing it, reflecting a possible training bias where accordions are commonly paired with human subjects. For the predicted class "Freight car," the heatmap indicates the model is misinterpreting part of a building structure as a freight car. Through example-based analysis, this confusion is better understood; there are visual similarities between building structures and freight car images in the dataset.

By using heatmaps to visualize the model's focus areas and then turning to example-based comparisons to investigate how training data may drive misclassification, this approach provides a clearer understanding of the model's interpretive challenges and highlights specific data biases. These insights are critical for guiding improvements to enhance model robustness and reduce error.

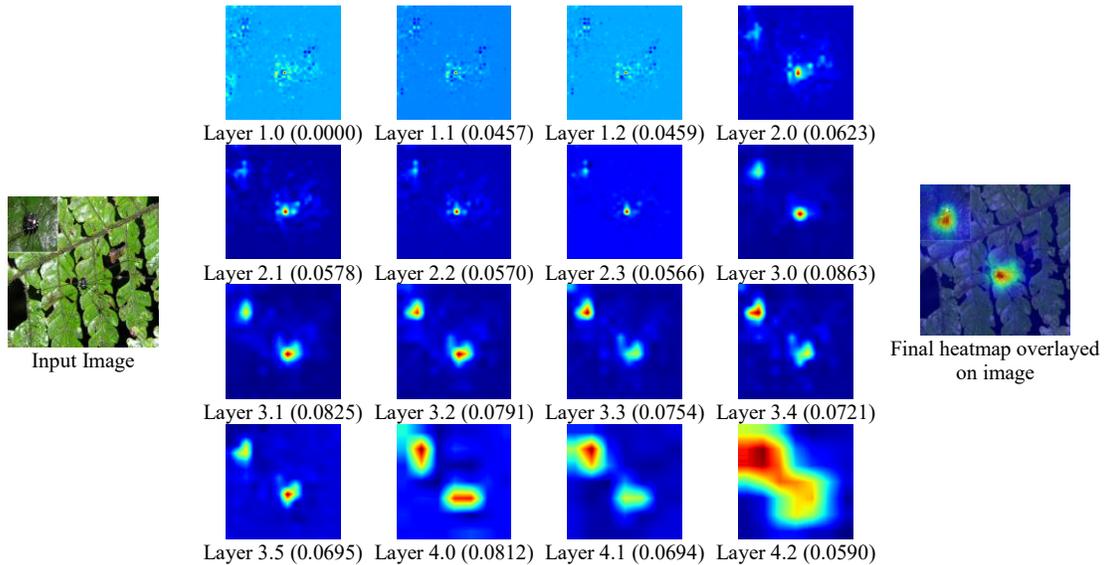

Figure 10: Heatmaps of different layers of ResNet along with their respective layer weights

### 3.1.4. Layer Weight Analysis

This section examines how integrating information across different layers affects visualizations and highlights the distinct roles various layers play in shaping the model's perception. Using the Integrative CAM (I-CAM) method, we generated heatmaps for layers within the ResNet-50 model, with each heatmap illustrating image regions that specific layers prioritize in classification decisions. The I-CAM method assigns adaptive weights to each layer's contribution, quantifying the importance of features extracted at different stages in the network.

By analyzing these weights, we can discern the influence of each layer on the model's overall decision-making process. For instance, comparing heatmaps across layers helps identify which layers focus on low-level, fine-grained features and which capture more abstract, high-level semantic information.

Fig. 10 illustrates this with an example image from the ImageNet validation dataset and its corresponding I-CAM heatmap for a ResNet-50 model pretrained on ImageNet-1K. Alongside the final I-CAM visualization, individual heatmaps for selected layers of ResNet-50 show that while the model's classification depends on the final convolutional layer (Layer 4.2), the model's perception is distributed across layers. Interestingly, the final layer is not the primary contributor to the model's interpretation of this image; instead, Layer 3.0, with a layer score of 0.0863, emerges as the most influential. Layer 4.2 ranks only as the 10th in importance, underscoring that earlier layers can significantly shape the model's perception even at the final decision stage.

This integrative approach to layer analysis provides deeper insights into how multi-layered feature fusion contributes to model interpretability, clarifying the roles of intermediate layers in supporting more robust, layered representations of visual data.

### 3.2. Quantitative Analysis using IoU Accuracy

The Integrative CAM (I-CAM) method, like other Class Activation Mapping (CAM) techniques, can be employed for weakly supervised localization. We quantitatively assessed I-CAM's localization effectiveness using the Intersection over Union (IoU) metric, which measures the alignment of the most salient heatmap regions with ground truth object locations. Specifically, IoU is calculated as the overlap between the predicted and ground truth areas, providing an objective measure of how well the heatmap highlights relevant regions that inform the model's decisions.

To perform this evaluation, heatmaps were generated for each test image using four visualization methods: Grad-CAM, Grad-CAM++, LayerCAM, and I-CAM. These heatmaps identify image regions deemed significant by the model for a given classification decision. Each heatmap was thresholded to create a localization mask, and the corresponding bounding box mask was then generated to calculate the IoU metric.

We calculated the average IoU for cases where the model's predicted class matched the true class, offering an overall measure of how often the most salient regions within the heatmaps aligned with ground truth bounding boxes. The formula for average IoU is as follows:

$$IoU_{avg} = \sum_k \frac{L_k^c \cap L_{Bbox}^c}{L_k^c \cup L_{Bbox}^c}$$

where $k$ is the number of correct predictions.

Table 1 presents the IoU evaluation results for the ImageNet dataset, showing that I-CAM outperforms other methods with superior localization accuracy. The findings further highlight I-CAM's utility in interpretability tasks and weakly supervised localization.

### 4. Conclusion

In this work, we introduced Integrative CAM (I-CAM), a method designed to deepen our understanding of how convolutional neural networks (CNNs) perceive and process images. I-CAM uniquely combines information across multiple layers, using adaptive fusion to provide a comprehensive view of model behavior. By integrating channel-wise biases and assigning scores to layers based on their relevance, I-CAM enhances traditional visualization methods with richer interpretability and offers insights beyond what single-layer or gradient-based techniques can provide.

A key advantage of I-CAM is its ability to bridge the gap left by conventional performance metrics, which can struggle to differentiate models with similar accuracy. When we use I-CAM to visualize feature extraction and representation capabilities, we gain a clearer perspective on which model may offer more reliable insights—a crucial factor when selecting models based not just on performance but also on interpretability.

Additionally, our generalization of the alpha term simplifies complex calculations involving higher-order derivatives. This can be especially valuable in other fields where such derivatives are difficult to handle, allowing for faster and more efficient computation. For cases where non-linear functions can be approximated as piecewise linear, this simplification remains beneficial, potentially impacting fields that rely on derivative-based methods.

The new layer scoring system introduced by I-CAM also gives us a more nuanced view of how different layers contribute to model interpretations. This scoring can guide model optimizations, such as simplifying less critical layers to make models more efficient for use in resource-limited environments. By going beyond just the final layer, I-CAM helps reveal the network's interpretive process across all layers, which can support better decision-making in models that depend on multi-layered information, ultimately improving classification accuracy.

We also found that combining I-CAM with example-based explanations sheds light on the reasons behind model misclassifications, reinforcing the importance of using multiple explainable AI (XAI) methods together. Just as having a range of tools allows for more thorough problem-solving, combining different XAI approaches provides a more complete understanding of model behavior. Identifying the best combination of XAI methods to achieve robust model interpretability is a promising avenue for future research. Furthermore, while we manually selected examples from the training dataset, future work could focus on automating this selection process using I-CAM itself to identify such examples [38].

One shared challenge with methods like I-CAM is maintaining visual accuracy when scaling heatmaps to match input image sizes. For example, in architectures like ResNet, the final convolutional layer has a resolution of $7x7$, while the input image is $224x224$. Upscaling the heatmap by a factor of 32 can introduce artifacts that may distort the visual representation. Addressing this limitation to retain visual fidelity at higher resolutions will be another important direction for future work.

Our current study also leaves open the question of whether layer prominence is specific to individual cases or generalized across classes. Further research into layer importance at the class level could improve I-CAM's adaptability to different tasks. The complexities of real-world datasets like ImageNet, with challenges such as class ambiguity, label inaccuracies, and multiple objects per image, result in data uncertainty in models and highlight the ongoing need for both better data quality and model robustness.

In terms of quantitative evaluation, we used Intersection over Union (IoU) to assess weakly supervised localization by comparing the I-CAM-generated object masks with ground truth bounding boxes. However, using bounding boxes rather than object masks is a compromise, as it can misrepresent localization accuracy. An annotated dataset with object masks for the ImageNet validation images would provide a more accurate basis for evaluating IoU and allow a fairer assessment of weakly supervised localization capabilities.

In summary, I-CAM represents a step forward in model interpretability, offering a powerful way to visualize and understand CNNs across all layers. By facilitating clearer model comparisons and supporting optimization, I-CAM can serve as a valuable tool for both researchers and practitioners. With continued improvements, particularly in areas like layer scoring and heatmap resolution, I-CAM has the potential to significantly contribute to transparency and reliability in deep learning.

**Acknowledgement**

We would like to express our sincere gratitude to DYSL-CT (DRDO), Chennai for funding the project (Contract No.: DYSL-CT/MMG/CARS/CS/23-24/01)